\def\year{2021}\relax
\documentclass[letterpaper]{article} 
\usepackage{aaai21}  
\usepackage{times}  
\usepackage{helvet} 
\usepackage{courier}  
\usepackage[hyphens]{url}  
\usepackage{graphicx} 
\usepackage{xcolor}
\usepackage{natbib}  
\usepackage{caption} 
\usepackage{hhline}
\usepackage{booktabs}
\usepackage{multirow}
\usepackage{subfigure}
\usepackage[linesnumbered,boxed,ruled,commentsnumbered]{algorithm2e}

\usepackage{color}

\title{DDRel: A New Dataset for Interpersonal Relation Classification in Dyadic Dialogues}
\author{
    Qi Jia,
    Hongru Huang,
    Kenny Q. Zhu
}
\affiliations{

	Shanghai Jiao Tong University\\
	Shanghai, China\\
	Jia\_qi@sjtu.edu.cn,
	onedesire@sjtu.edu.cn,
	kzhu@cs.sjtu.edu.cn
}

\begin{document}

\maketitle 

\begin{abstract}
Interpersonal language style shifting in dialogues is an interesting and almost instinctive ability of human. Understanding interpersonal relationship from language content is also a crucial step toward further understanding dialogues. 
Previous work mainly focuses on relation extraction between 
named entities in texts.  In this paper, we propose the task of 
relation classification of interlocutors based on their dialogues.  
We crawled movie scripts from IMSDb, and annotated the relation labels for each session according to 13 pre-defined relationships. 
The annotated dataset DDRel consists of 6300 dyadic dialogue sessions 
between 694 pair of speakers with 53,126 utterances in total. 
We also construct session-level and pair-level relation classification tasks with widely-accepted baselines. The experimental results show that 
this task is challenging for existing models and the dataset will be useful 
for future research.
\end{abstract}

\section{Introduction}

Interpersonal relationship is an implicit but important feature 
underlying all dialogues, shaping how language is used and perceived 
during communication. People start to practice such style shifting in 
communication at very early stage unconsciously. Study~\cite{mind-reading} 
finds that when children listen to another person, their understanding 
of the counterpart depends on nature of their relationship with the speaker. 
Study~\cite{conversational-motive} also find that conversations between 
different partners are executed under different interpersonal motives, 
and thus the dialogues differ in topics and styles. 
Also, similar expressions may reflect different emotions and attitudes in 
different relationships. 

Analyzing the relationship based on dialogues between interlocutors 
is well-motivated. First, it can provide dialogue systems with 
supplementary features for generating more suitable responses for different 
relationships, which helps in developing more intelligent role-playing chatbots.
Second, it is useful in recommendation systems if the system can figure 
out the relationship between users according to their privacy-insensitive chats.
Third, an automatic relationship classifier can help 
understand where the interpersonal stress/stimuli comes from in 
mental disorder treatment~\cite{tension-monitor,bopolar-monitor,cog-load}.
Besides, it can be also used for crime investigation, relieving
the burden of manual monitoring and improve the productivity of searching
in large amount of dialogue data. 

\begin{figure}[t!]
	\centering
	\includegraphics[width=0.85\columnwidth]{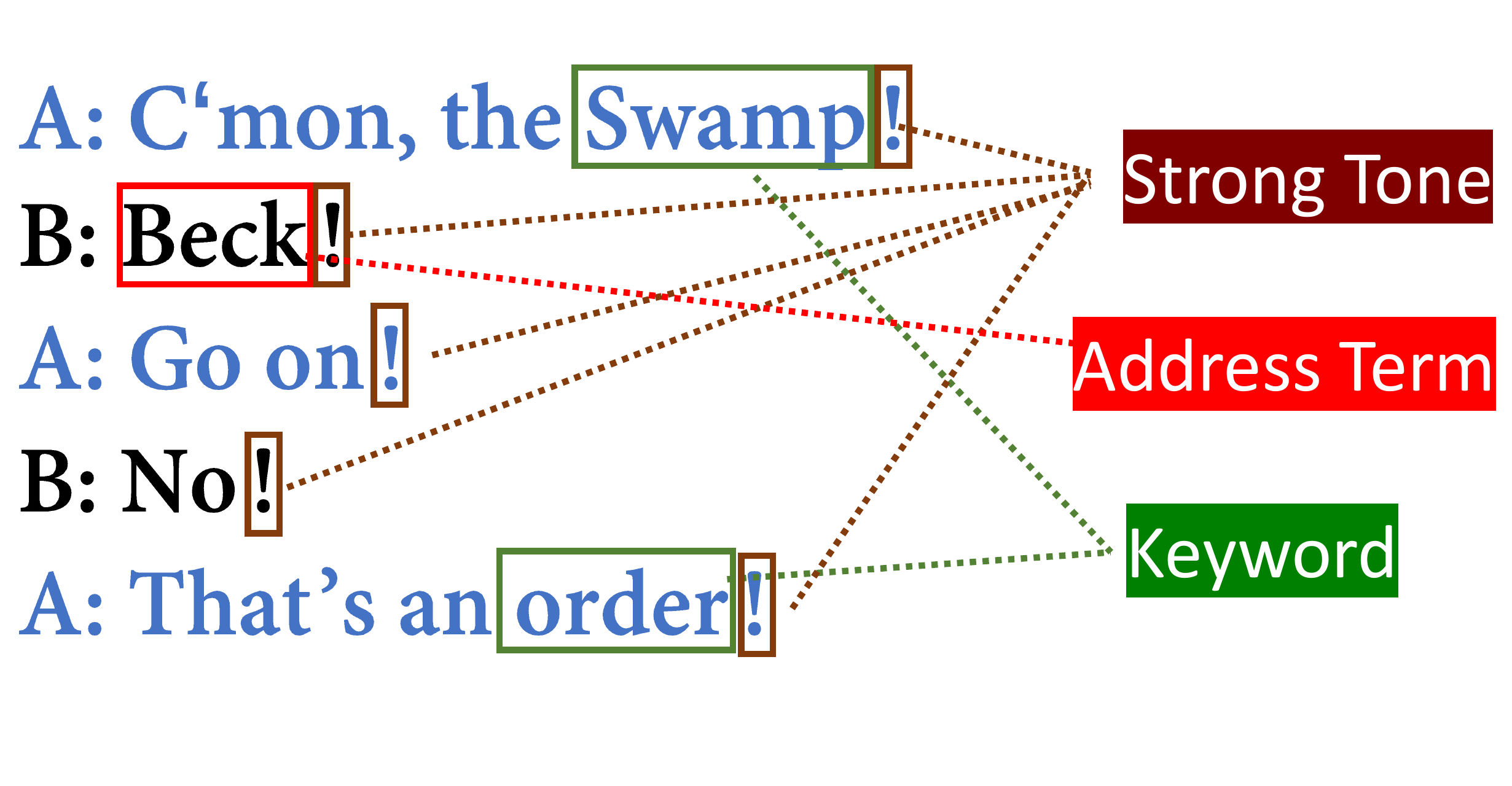}
	\caption{A sample dialogue and parts that are 
		reported as informative by human testers. }
	\label{fig:instinct}
\end{figure}

Relation classification of dialogue sessions is not an easy task. Figure \ref{fig:instinct} shows a 5-turn dialogue example. We can see that it's challenging to fully contextualize such short conversation without any prior knowledge, 
except one might infer that the two speakers are fellow soldiers 
in the military. Facing such problem, human usually resort to their 
communication experiences and commonsense knowledge to make sense of the background stories and give the inferences about the speakers' relationship. This shows that the inference of relationships from dialogues is possible but not straight forward for statistical models.

Previous work mainly focuses on relation classification between named entities. 
Sentence-level relation classification datasets, such as FewRel~\cite{HanZYWYLS18} and  TACRED~\cite{ZhangZCAM17}, have been widely studied~\cite{Zhang0M18,GaoH0S19,ZhangHLJSL19}, targeting on figuring out the correct relation type between entities within a given sentence. Recent research sets sights on the 
inter-sentence relations. DocRED~\cite{YaoYLHLLLHZS19} has been proposed 
as the largest document-level relation extraction dataset from plain text, 
where person (an entity type) only occupies 18.5\% of the entities. 
DialogRE~\cite{YuSCY20} aims at predicting the relations between two arguments 
in dialogues and MPDD~\cite{ChenHC20} is a Chinese dataset for predicting relations between speaker and listeners of each turn in a dialogue session.
However, relations in both datasets are limited to the current dialogue session, without cross-session considerations.

Different from their problem definition, we are trying to figure out the interpersonal 
relationships between speakers in dyadic dialogues from two perspectives, session-level and pair-level.
Multiple dialogue sessions may happen between each pair of speakers and 
it is usually not easy to figure out the relationship between 
a pair of speakers with only one session and no background context. 
Human has the ability to make the connections between multiple sessions 
and construct the whole picture between two speakers. 
In other words, cross-session inferences are required for final predictions. 

In this paper, we propose a new dyadic dialogue dataset for 
interpersonal relation classification called DDRel. 
The dataset consists of 6300 dialogue sessions from movie scripts crawled 
from IMSDb between 694 pair of speakers, annotated with relationship labels 
by human. 13 relation types according to 
Reis and Sprecher~\cite{reis2009encyclopedia} are covered in our dataset and these types can cover most of the interpersonal relations in daily life. 
Several strong baselines and human evaluations are implemented. 
The results and future work of our dataset are discussed.

In summary, this paper makes following contributions:
\begin{itemize}
	\item We propose the task of dialogue relation classification for speakers, different from the previous intra-sentence or inter-sentence relation classification tasks (Sec.~\ref{sec:task}).
	\item To the best of our knowledge, we construct the first-ever dialogue relation classification dataset for analyzing interpersonal relationships between speakers with multiple dialogue sessions (Sec.~\ref{sec:dataset}). 
	\item We establish a set of classification baselines on our dataset using standard learning-based techniques. The gap between SOTA models and human performances show the difficulty of this task and higher requirements for current models (Sec.~\ref{experiments} and Sec.~\ref{sec:results}).
\end{itemize}

\section{Related Work}

\subsection{Relation Classification}
Relation classification or extraction is an important first step for constructing structured knowledge graph in NLP with the popular benchmark datasets such as NTY-10~\cite{RiedelYM10} and the SemEval-2010 dataset~\cite{HendrickxKKNSPP10}. Previous datasets for relation classification focus on figuring out the relation type between two entities in a single sentence, including FewRel~\cite{HanZYWYLS18} and  TACRED~\cite{ZhangZCAM17}.  However, such intra-sentence relation classification has a limitation in real applications and looses nearly 40.7\% of relational facts according to previous research~\cite{SwampillaiS10,VergaSM18,YaoYLHLLLHZS19}.

Inter-sentence relation classification or document-level relation classification has gained more attention in recent years. There are only several small-sized dataset for this task, including a specific-domain dataset PubMed~\cite{LiSJSWLDMWL16} and two distant supervised datasets from Quirk and Poon~\shortcite{QuirkP17} and Peng et al. ~\shortcite{PengPQTY17}. To facilitate the research in this area, DocRED~\cite{YaoYLHLLLHZS19} has been proposed as the largest dataset for document-level relation classification. Our task is different from it since we focus on interpersonal relations while person-related entities is only a small component in DocRED. Besides, our task is based on dialogue sessions instead of plain documents and interpersonal relation classification may need inferences beyond session level.

\subsection{Dialogue Datasets}
Dialogue system is a hot research point in recent years with a rapid growing number of available dialogue datasets. 

Generally, dialogue datasets can be divided into two categories. 
One is the task-oriented dialogue datasets such as Movie Booking Dataset~\cite{LiCLGC17}, CamRest676~\cite{UltesRSVKCBMWGY17} and MultiWOZ~\cite{BudzianowskiWTC18}. These datasets focus on single or multiple targeting domains and are usually labeled with dialogue act information, serving for the slot filling~\cite{LiuWXF20} and dialogue management tasks~\cite{BudzianowskiV19} when building task-oriented dialogue systems.
The other is the open-domain chit-chat datasets such as DailyDialog~\cite{LiSSLCN17}, MELD~\cite{PoriaHMNCM19} and PERSONA-CHAT~\cite{KielaWZDUS18}. The resource of these conversations are usually social media platforms, including Facebook, Twitter, Youtube, and Reddit. Researches on these datasets mainly focus on emotion recognition and emotion interplay among interlocutors, helping chatbots generate more emotionally coherent~\cite{GhosalMPCG19} and persona consistent responses~\cite{ZhengZHM20}.

There are two existing datasets similar to our settings. One dataset is the DialogRE~\cite{YaoYLHLLLHZS19}. It focuses on predicting the relations between two arguments in a dialogue session, where relations between arguments of interlocutors are rare. Also, since all of the 1,788 dialogue sessions are crawled from the transcript of \textit{Friends}, it suffers a limitation of the diversity of scenarios and speakers. Another dataset is MPDD~\cite{ChenHC20}. This dataset contains 4,142 dialogues annotated with speaker-listener interpersonal relation in multi-party dialogues for each utterance, while the relation types in our dataset is not such directional relationships. Besides, both datasets ignore the fact that interlocutors may have multiple sessions which is considered in our task and dataset. Our task is more reasonable with practical social meanings.

\section{Task Definition}
\label{sec:task}

Our work aims at identifying the interpersonal relation between interlocutors in dyadic dialogues. 
The types of relationships are pre-defined, annotated as $R=\{R_1, R_2, ..., R_m\}$ where $m$ is the number of relation types.
A number of sessions may happen between the same pair of interlocutor. So, we define the relation classification task in two levels: session-level and pair-level.

Given the $j$-th dialogue session $D_j^i$ between the $i$-th pair of interlocutors,  \textbf{session-level relation classification task} is to inference the most possible relation type for this session, i.e.:
\begin{equation}
R_j^i = \arg\max_{R} f_s(D_j^i)
\end{equation}

Due to the fact that it's quite hard for even human to fabricate the whole story only through one dialogue session, \textbf{pair-level relation classification task} is defined as follows. Given the dialogues between the $i$-th pair of interlocutors denoted as $D^i=(D_1, D_2, ..., D_n)$, pair-level relation classification task is to figure out the most possible relation type for this pair, i.e.:
\begin{equation}
R^i = \arg\max_{R}f_p(D_i) = \arg\max_{R}f_p(D_1, D_2, ..., D_n)
\end{equation}

13-class taxonomy of relationships are covered in our DDRel dataset, 
including {\em child-parent}, {\em child-other family elder}, {\em siblings}, {\em spouse}, {\em lovers}, {\em courtship}, {\em friends}, {\em neighbors}, {\em roommates}, 
{\em workplace superior-subordinate}, {\em colleagues}, {\em opponents} and 
{\em professional contacts},
based on Reis and Sprecher~\cite{reis2009encyclopedia}, 
in which they elaborate on psychological 
and social aspects of various relationships. 
We define these categories by social connections because they make general 
sense in life. Although individual difference exists 
in every real-world case, it was found that such relationship category 
has different expectations, 
special properties (e.g., marriage usually involves sex, 
shared assets and raising children)~\cite{argyle1983sources}, 
distinctive activities (e.g., talking, eating, drinking and 
joint leisure for friendship) and 
rules~\cite{argyle1984rules} of its own, which are agreed across cultures.
Note that this is not an all-round coverage of 
all possible relationships in human society 
and we aim to cover those common
ones in real life which may be of interest in interpersonal 
relationship research.
These fine-grained labels are prepared for possible related future research.

To evaluate the classification ability of the model from coarse-grained to fine-grained, we also clustered our 13 specific 
relations types into 6 classes and 4 classes considering the social field, the seniority 
and the closeness between two speakers. The details of relation types are listed
in Table \ref{table:relationtypes}.

\section{Dataset}
\label{sec:dataset}
Although there are many currently available dialogue datasets, 
most of them are used for training automatic dialogue robots/systems, 
thus they either do not cover the diversity of interpersonal relationships, 
or do not come with relationship labels.
Therefore, we build a new dataset 
composed of $6,300$ sessions of dyadic dialogues with 
interpersonal relationship labels between two speakers, 
extracted from movie scripts crawled from 
the Internet Movie Script Database (IMSDb). More details are explained as follows. 

\begin{table*}[t]
	\centering
	\small
	\begin{tabular}{@{}lllrrrrrr@{}}
		\toprule[1.5pt]
		\textbf{4 classes} & \textbf{6 classes} & \textbf{13 classes}  & \textbf{\# Sessions} & \textbf{\% Sessions}& \textbf{\# Pairs} & \textbf{\% Pairs} & \textbf{\# Turns} & \textbf{\% Turns}  \\ 
		\hline
		\multirow{4}{*}{Family}&\multirow{2}{*}{Elder-Junior} & Child-Parent   &  414    &   6.57   &  67   &	9.65	&  3,377 & 6.36   \\
		& & Child-Other Family Elder 															    &   91   &  1.44    &  12   &	1.73	&   632  & 1.19  \\
		& \multirow{2}{*}{Family Peer} & Siblings 														  &   211   &   3.35   &  27   &   3.89	&   1,585  & 2.98 \\
		& & Spouse 																						   &   568   &   9.02   &  51   &	 7.34  &   4,784  & 9.01 \\
		\hline
		\multirow{2}{*}{Intimacy}& \multirow{2}{*}{Intimacy} &  Lovers					&  1,852    &  29.40    &  244   &20.75	&   17,474 & 32.89  \\
		& & Courtship	
		&  146    &   2.32   &  15   & 2.16	&  1,323 & 2.49   \\
		\hline
		\multirow{3}{*}{Others}& \multirow{3}{*}{Peer} & Friends 					&  1,049    & 16.65     &  124   &17.87	&  8,900 & 16.75    \\
		& &Neighbors 																					 &   21   &  0.33    &  2   &	0.29& 189 & 0.36      \\
		& &Roommates 																				    &  120    &   1.90   &  8   & 1.15	&   966 & 1.82   \\

		\hline
		\multirow{4}{*}{Official}& \multirow{1}{*}{Elder-Junior} &Workplace Superior-Subordinate &  536    &   8.51   &  79   &	11.38& 3,958 & 7.45     \\
		&\multirow{3}{*}{Official Peer} &Colleague/Partners										&   710   &  11.27    &  76   &	10.95 & 5,455 & 10.27     \\
		& &Opponents																				  &  203   &  3.22    &  33   &  4.76	&  1,532 & 2.88   \\
		& &Professional Contact																	  &   56   &  8.07    &  56   &	8.07 & 2,952 & 5.56     \\
		
		\bottomrule[1.5pt]
		
	\end{tabular}
	\caption{Statistics on categories of interpersonal relation types.}
	\label{table:relationtypes}
\end{table*}

\subsection{Dataset Extraction and Processing}
Initially, we crawl 995 movie scripts from IMSDb, and 941 of them remain after we automatically match the titles with movies in IMDb\footnote{\url{https://www.imdb.com/}} and filter out those that do not meet following requirements:
\begin{itemize}
	\item Don't have a match in IMDb; 
	\item Not in English; 
	\item Very unpopular(measured by number of raters). 
\end{itemize}

By observing the formats of the scripts and manually defining text patterns, 
we split each script into scenes, extract the sequence of (speaker, utterance) 
pairs for each scene and identify subsequences that meet the following requirements  
as dyadic dialogue sessions: 
\begin{itemize}
	\item Two speakers speak alternately without being interrupted by a third one; 
	\item Each dialogue session contains at least 3 turns. 
\end{itemize}

We set this minimum length requirement to make sure that two speakers are speaking to each other instead of participating in a group discussion. Finally, we count the total number of turns taken between each pairs and filter out those having fewer than 20 turns to make sure the relationship between the two speakers is significant and not as trivial as greetings between strangers. This filtering step also helps reduce the cost of labeling because more sessions can share the same pair of speakers.

\subsection{Annotation Procedure}
Although interpersonal relationships are not static or mutual exclusive, most of them exhibit relative stability over time~\cite{gadde1987stability}, and relationships in movies are usually more clear-cut. Therefore, in this paper, we model relationship as a single, stable label. Such assumption simplifies our task and significantly reduces the workload of labeling, though it introduces 
ambiguity in certain cases such as evolving relationships (e.g., courtship $\rightarrow$ lover $\rightarrow$ spouse) or concurrent ones that do not usually exist together(e.g., enemies falling in love). To avoid these situations, we require the annotator to only assign labels when the relationship is clear, relatively stable and typical.

Our ground truth annotator was provided with the movie title, the pair of characters involved in the dialogue, movie synopsis from IMDb and Wikipedia for each movie, as well as complete access to the Internet, and was asked to choose between one out of tens of classes mentioned in Reis and Sprecher's work~\cite{reis2009encyclopedia} or ``Not applicable (NA)'' label.
It took the annotator 100 hours across one and a half months to finish the annotation of 300 movies, at a rate of approximately 4.07 minutes per pair. Only 47.11\% of the pairs received a specific label, while others are considered ``not applicable''. Finally, 13 kinds of relation types are labeled in our dataset, covering a variety of interpersonal relationships and enough for developing classification methods on this task.

\textbf{Second-Annotator Verification}
Due to excessive cost of the annotation task, we are not able to commit multiple annotators on the labeling task. But to compensate that, we verify the accuracy of annotation by having a second person label 100 pairs with the same experimental settings. The inter-annotator agreement (kappa) is 82.3\% for 13-classes. This indicates that incorrect labels are limited, and the annotation by the first human is reliable.

\subsection{Dataset Statistics}
The current version of the DDRel dataset~\footnote{We have processed $941$ scripts and 
	manually labeled $300$ of them with relationships at present.}
contains $6,300$ labelled sessions of dyadic dialogues, taking place between $694$ pairs of interlocutors across $300$ movies. 
The average number of turns in each dialogue is $8.43$, 
while it varies greatly (the standard deviation is $6.94$).  The number of sessions for each pair of interlocutors also varies a lot with $avg=9.08$ and $std=7.80$. The whole dataset is split into train/development/test sets by 8:1:1 as shown in Table \ref{table:dataset}. All of the dialogue sessions between the same interlocutors are assigned to the same subset and there is no overlap between three subsets.

The distribution of the whole dataset on 13 relation types are shown in Table \ref{table:relationtypes}. {\em Lovers}, {\em Friends} and {\em Colleague/Partners} are the three largest classes and take up about half of the dataset, while the smallest relation type {\em Neighbor} only have 2 pairs of interlocutors with 21 dialogue sessions. The proportion of different relation types are unbalanced, aggravating the difficulty of classification tasks.

\section{Experiments}
\label{experiments}

In this section, we introduce the baseline models, human evaluation settings and evaluation metrics.

\subsection{Baseline Models}
\label{sec:baselines}
We introduce two naive baseline methods, Random and Majority, and three strong neural
baseline models, CNN, LSTM and BERT. 
The code and dataset are avaliable at Github~\footnote{\url{https://github.com/JiaQiSJTU/DialogueRelationClassification}}.

\textbf{Random:} A relation type is randomly assigned to a dialogue session or a pair of interlocutors.

\textbf{Majority:} The most frequent relation type is assigned to a dialogue session or a pair of interlocutors.

\begin{table}[h!]
	\centering
	\small
	\begin{tabular}{@{}lrrr@{}}
		\toprule[1.5pt]
		\textbf{}                				& \textbf{train} & \textbf{development} & \textbf{test} \\ 
		\hline
		\# Pair of Speakers      		  &   541     &  75     &   78  \\
		\# Sessions             			 &     5,037   &    653    &  610  \\
		\# Turns                			   &   42,564     &    5,210    &  5,352  \\
		Sessions per pair (mean)  	  &    9.31    &   8.71     &  7.82  \\
		Sessions per pair (std)        &    8.18    &   6.35     & 5.96   \\
		Turns per session (mean)    &    8.45     &   7.98     &   8.77 \\
		Turns per session (std)      &    6.96     &   5.60     &  7.93  \\
		\bottomrule[1.5pt]
		
	\end{tabular}
	\caption{Statistics on the splitted datasets.}\label{table:dataset}
\end{table}

\begin{table*}[th]
	\centering
	\small
	\begin{tabular}{@{}llcccccc@{}}
		\toprule[1.5pt]
		&               				& \multicolumn{2}{c}{\textbf{4-class}} & \multicolumn{2}{c}{\textbf{6-class}} & \multicolumn{2}{c}{\textbf{13-class}} \\ 
		& & Acc & F1-macro                 & Acc & F1-macro                  & Acc & F1-macro \\
		\midrule
		\multirow{6}{*}{\textbf{Session-level}}&Random   &23.0$\pm$3.56 &22.67$\pm$3.71 &17.33$\pm$2.62 & 15.80$\pm$3.00& 8.33$\pm$2.62& 6.63$\pm$2.12  \\		
		&Majority    &31.00$\pm$0.00 &11.80$\pm$0.00 &31.00$\pm$0.00 &7.90$\pm$0.00 &26.00$\pm$0.00 &3.20$\pm$0.00 \\
		&LSTM    &29.80$\pm$1.28 &22.87$\pm$1.24 &30.83$\pm$1.16 &11.10$\pm$0.08 &28.50$\pm$1.44 &4.63$\pm$0.45 \\
		&CNN    &42.67$\pm$2.93 & 33.27$\pm$6.63&37.80$\pm$1.31 & 31.40$\pm$6.67 &32.33$\pm$2.46 &9.20$\pm$4.97 \\
		&BERT   &47.10$\pm$1.28 &44.53$\pm$1.10 &41.87$\pm$0.81 &39.40$\pm$0.85 &39.40$\pm$0.36 &20.40$\pm$0.67 \\
		&Human &56.00$\pm$6.00 &55.20$\pm$6.30&50.00$\pm$9.00&53.00$\pm$8.10&38.50$\pm$5.50&40.75$\pm$8.15 \\ 
		\midrule
		\multirow{6}{*}{\textbf{Pair-level}}&Random   &28.20$\pm$9.30 &26.90$\pm$9.24 &17.93$\pm$7.89 &16.2$\pm$7.54 &6.43$\pm$2.76 & 5.73$\pm$2.64 \\		
		&Majority    &23.10$\pm$0.00 &9.40$\pm$0.00 &23.10$\pm$0.00 &6.20$\pm$0.00 &19.20$\pm$0.00 & 2.50$\pm$0.00\\
		&LSTM    &25.63$\pm$2.76 &13.13$\pm$5.06 &22.67$\pm$0.61 &6.40$\pm$0.29 &19.20$\pm$0.00 &2.57$\pm$0.05 \\	
		&CNN    &47.47$\pm$2.76 &35.03$\pm$5.80 &38.47$\pm$4.21 &30.40$\pm$9.06 & 22.20$\pm$6.08& 7.07$\pm$6.04\\	
		&BERT   &58.13$\pm$0.61 &52.00$\pm$0.86 & 42.33$\pm$2.76&38.00$\pm$1.14 &39.73$\pm$1.79 &24.07$\pm$0.63 \\
		&Human & 75.65$\pm$3.85 &73.00$\pm$4.40 & 72.40$\pm$4.50&73.55$\pm$5.45 &63.45$\pm$1.95 &54.40$\pm$3.00 \\ 
		\bottomrule[1.5pt]
		
	\end{tabular}
	\caption{The classification results(\%) on session-level tasks and pair-level tasks.}
	\label{tab:results}
\end{table*}

\textbf{CNN:}
TextCNN, proposed by Kim ~\shortcite{Kim14}, is a strong text classification models 
based on convolution neural network.
All of the utterances in a dialogue session are concatenated as the input to the 
embedding layer, where 300-dimension pre-trained Glove~\shortcite{pennington2014glove} 
embeddings are used and freezed 
during training. Following the setting of Kim ~\shortcite{Kim14}, we used three convolution layers
with kernel size 3, 4, and 5 to extract semantic information from the dialogue. A dropout layer with 
probability $0.5$ is attached to each convolutional layer to prevent overfitting.
Finally, a linear layer and a softmax function are set for the final predictions. 
The loss function is the negative log likelihood loss. Stochastic gradient descent is 
used for parameter optimization with the learning rate equaling $0.01$. 

\textbf{LSTM:}
The attention-based bidirectional LSTM network by Zhou et al.~\shortcite{ZhouSTQLHX16} 
is implemented as another neural baseline. The same pre-trained Glove embeddings are 
used for the embedding layer. Then high-level features are extracted by a single 
Bi-LSTM layer. The last hidden states of both directions are concatenated as the 
query to do the self-attention among the input words. Finally, the weighted summed 
feature vector can be used to characterize the whole session and used for final relation 
classification with a linear layer and a softmax function. We use AdamDelta as optimizer with
learning rate $0.0003$ following Zhou et al.~\shortcite{ZhouSTQLHX16}.

\textbf{BERT:}
We fine-tuning the uncased base model of BERT released by 
Devlin et al.~\shortcite{DevlinCLT19}. 
All of the utterances in a dialogue session 
are also concatenated with the special token [CLS] as the start of the sequence.
Following the general procedure of fine-tuning BERT, we pass the output hidden state of 
[CLS] token into a fully connected layer for classification and use Adam as optimizer
with learning rate $1e-6$. We fine-tune the dataset for 32 epochs with early stopping
patience equal to 3.

The above baselines can be directly used for session-level classifications. For pair-level classifications, we do the following calculation for each neural baseline based on the session-level trained models: We first calculate the MRR metric of each relation type for each dialogue session. Then, given a pair of interlocutors with multiple sessions, the confidence score for each relation type can be regarded as the average MRR among sessions. Finally, the relation type with the maximum confidence score is regarded as the final prediction for this pair.

\subsection{Human Evaluation Settings}

To give an upper bound of our proposed DDRel dataset, we hired human annotators to do the relation classification tasks on the test set. Since given the 13-class classification results, the 4-class or 6-class classification results are obvious for human. We only asked annotators to do the 13-class interpersonal relation classification tasks.

We asked 2 volunteers to do the 13-class relationship task on session-level samples.  Each session are showed individually and volunteers are required to choose the most possible relation type. 
Another 2 volunteers are hired to do the 13-class relationship task on pair-level samples. All of the dialogue sessions between a pair of speakers are given to the volunteers to inference the relation types.

\subsection{Evaluation Metrics}
\label{sec:metrics}
Each classification model are trained separately for relation classification tasks on different granularity. We use accuracy and F1-macro scores for evaluation.

\section{Results and Discussions}
\label{sec:results}

\begin{figure*}
	\centering
	\textbf{\em  4-class pair-level results. \quad\quad \quad            6-class pair-level results.    \quad      \quad\quad    13-class pair-level results}\vspace{-1.0cm}\\
	\subfigure{
		\begin{minipage}{0.25\linewidth}
			\includegraphics[width=1\linewidth]{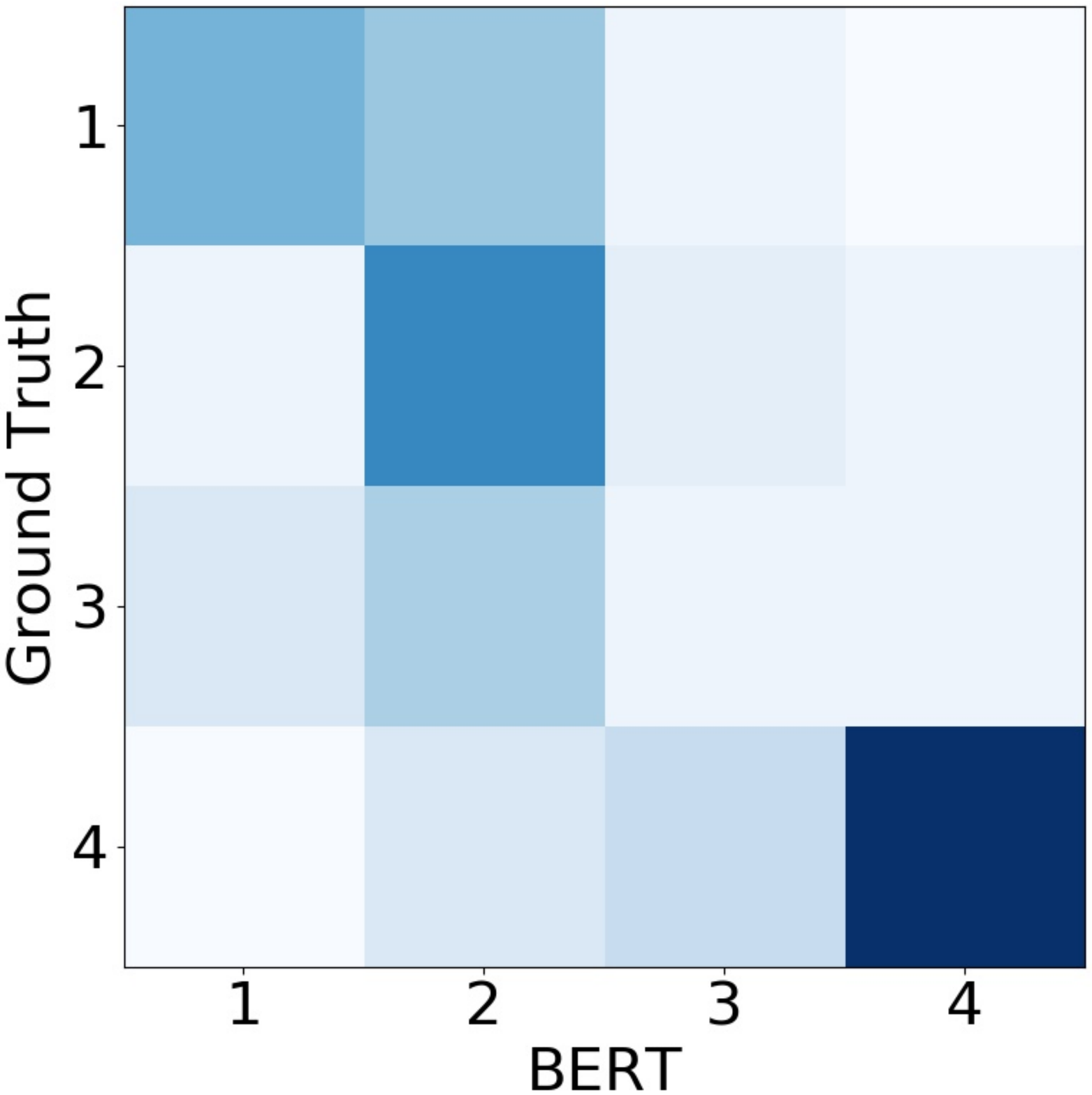}\vspace{-1.75cm}
			\includegraphics[width=1\linewidth]{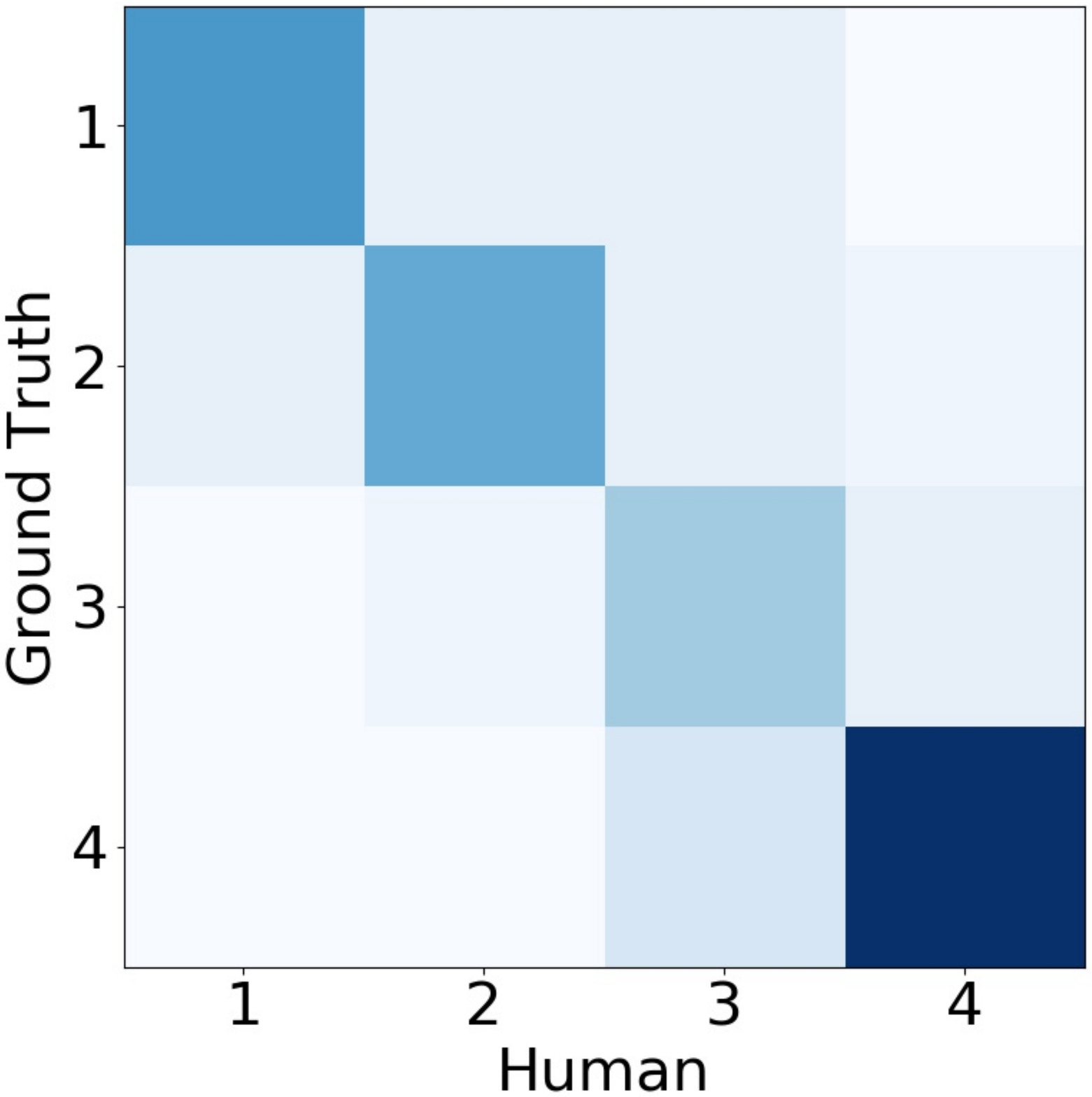}
	\end{minipage}}
	\subfigure{
		\begin{minipage}{0.25\linewidth}
			\includegraphics[width=1\linewidth]{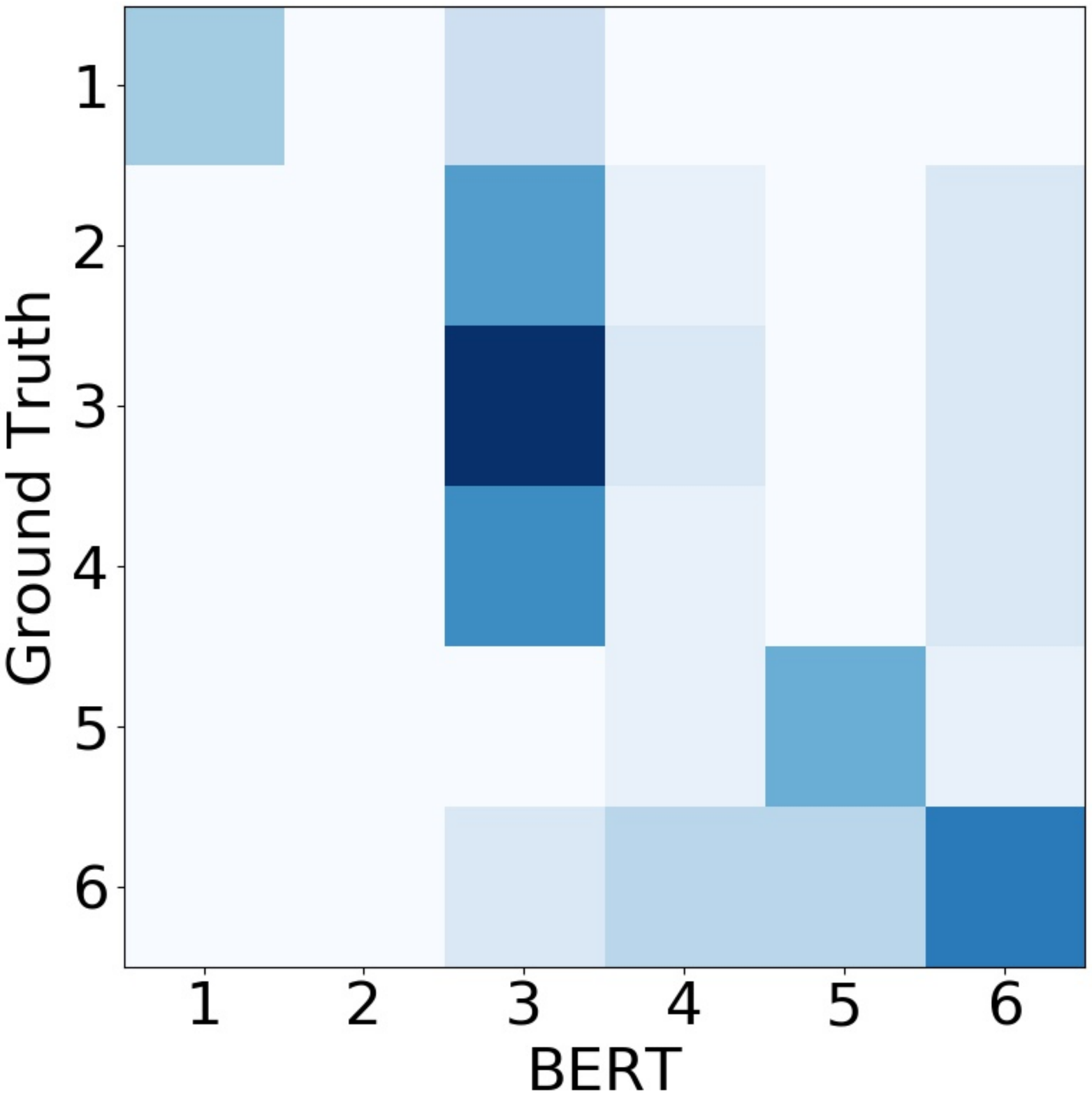}\vspace{-1.75cm}
			\includegraphics[width=1\linewidth]{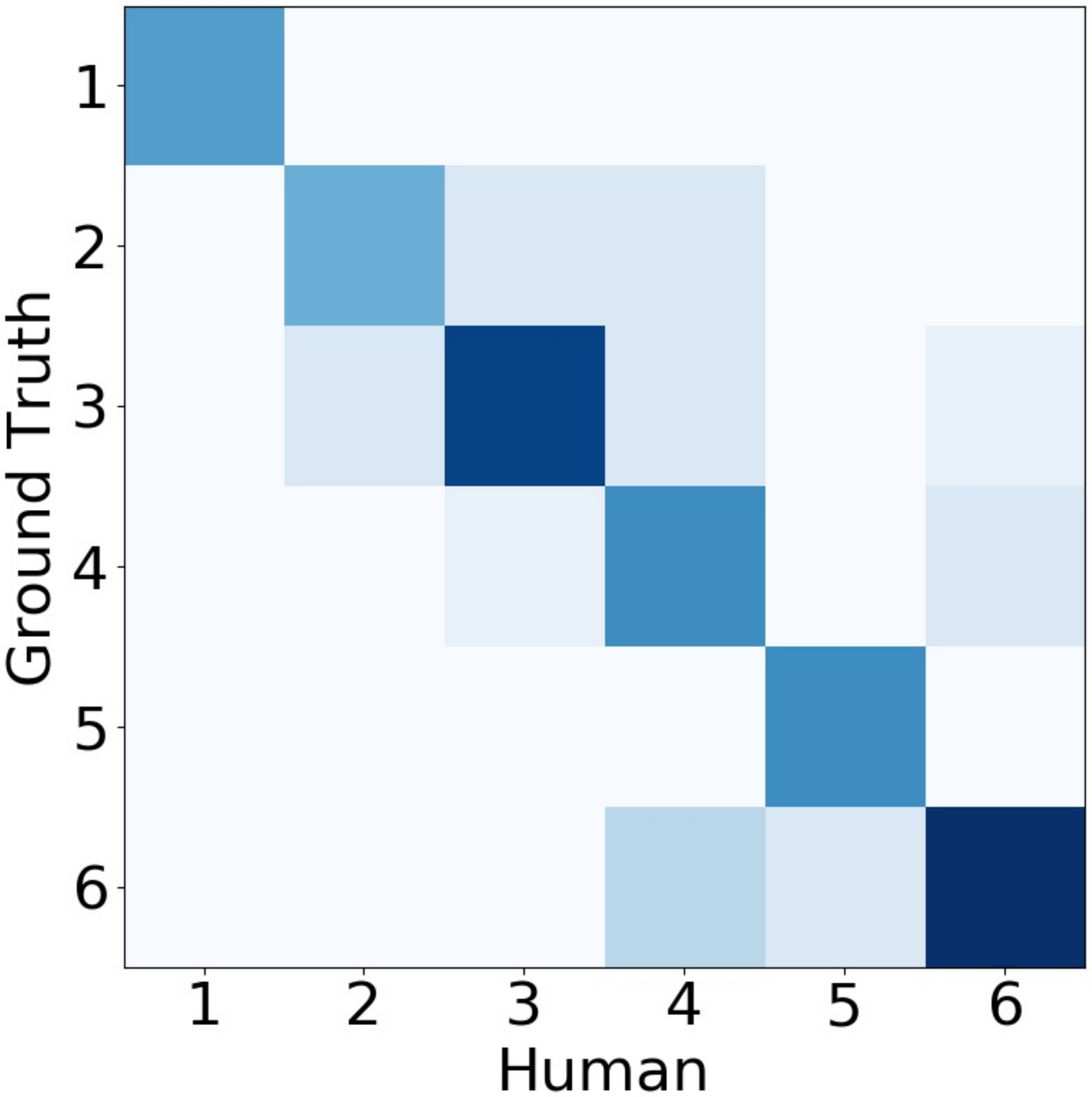}
	\end{minipage}}
	\subfigure{
		\begin{minipage}{0.25\linewidth}
			\includegraphics[width=1\linewidth]{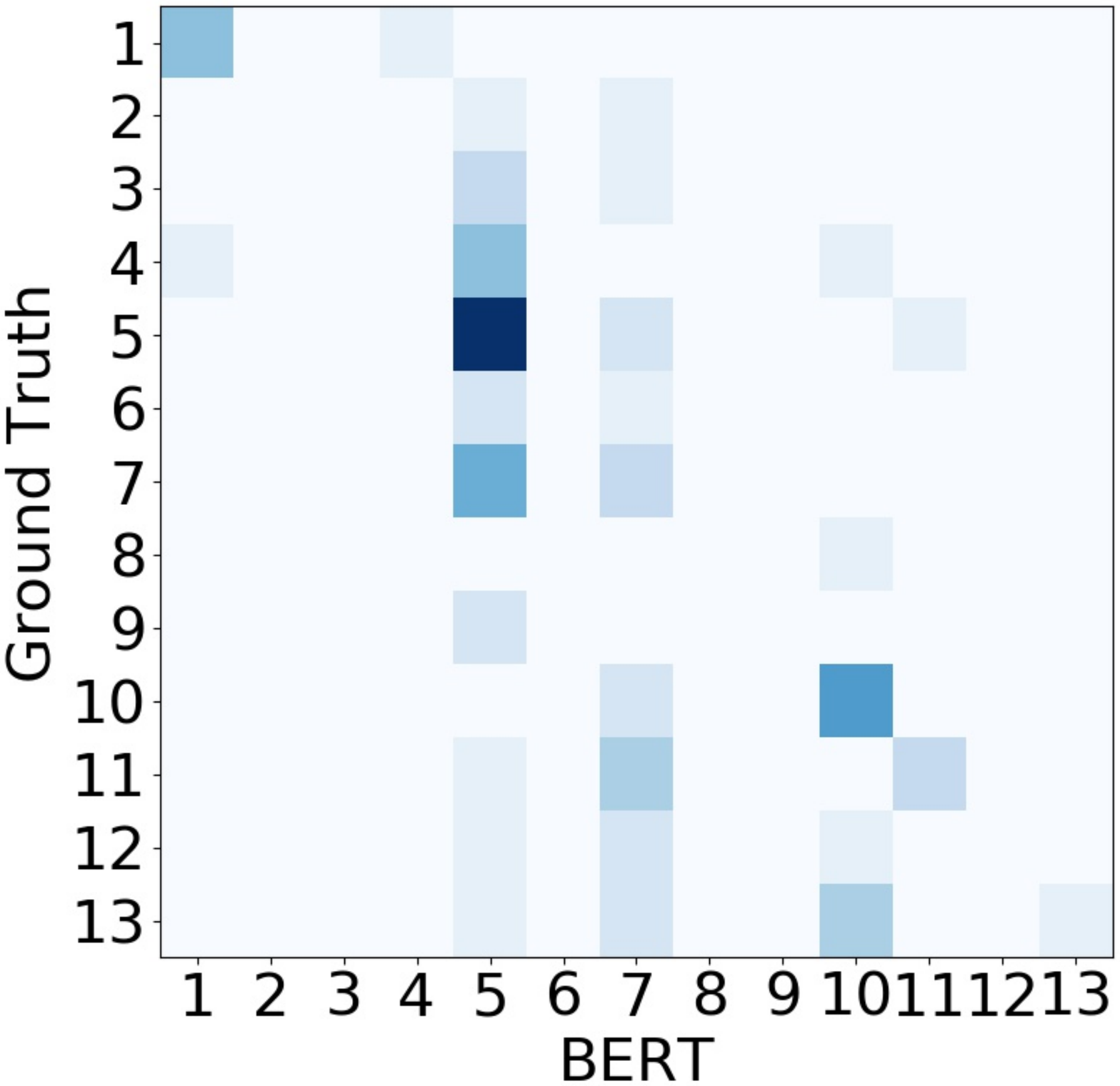}\vspace{-1.75cm}
			\includegraphics[width=1\linewidth]{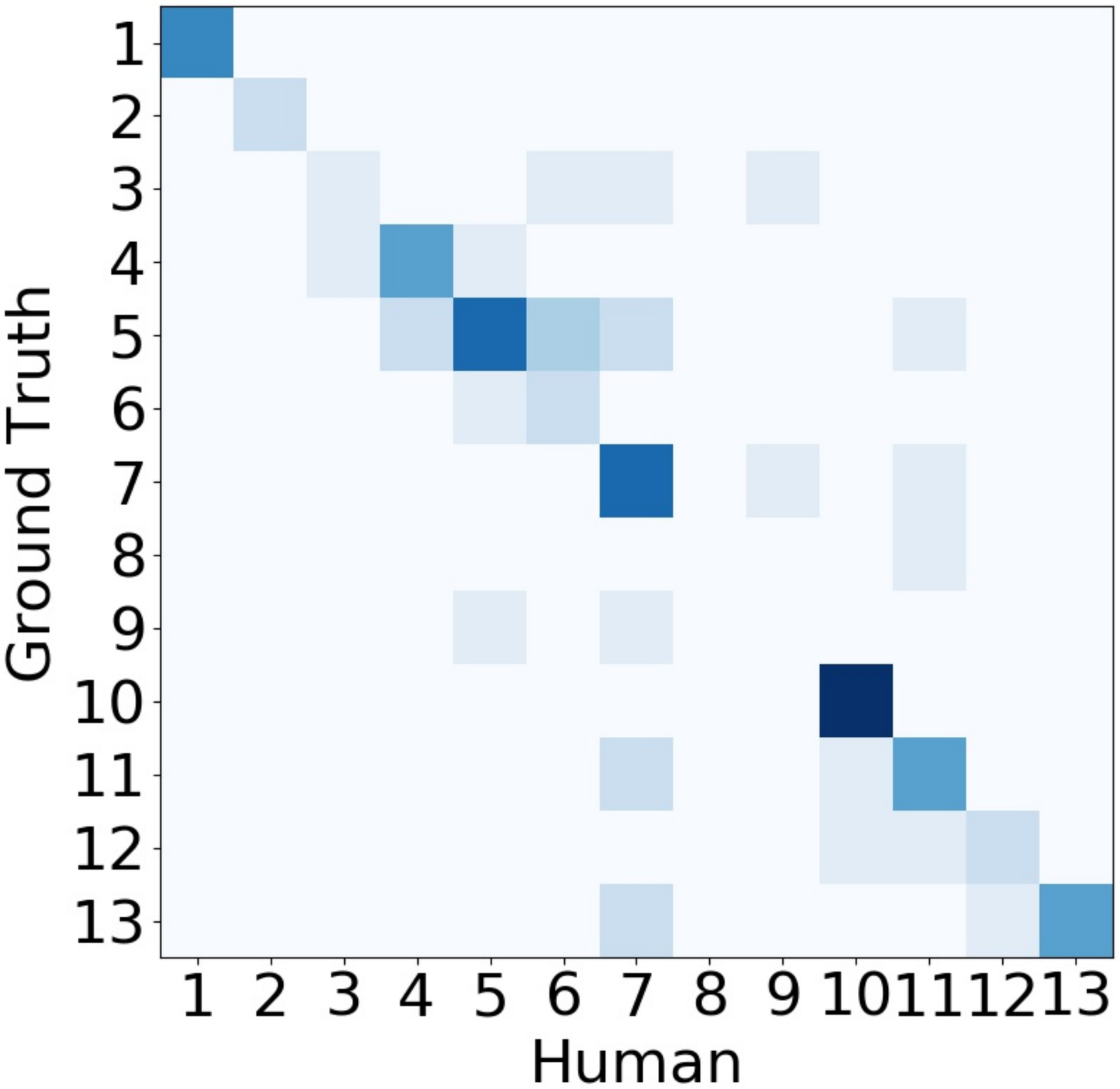}
	\end{minipage}}
	\vspace{-1cm}
	\caption{The confusion matrix of relation classification tasks.}
	\label{fig:confusion}
\end{figure*}
In this section, we discussed the classification performances and a simple data augmentation method for pair-level classification with a case study and future directions.
\subsection{Session-level Performance}
We run all of the baseline models with three different random seeds and then obtain the mean and
std value of the evaluation metrics. The results of baselines and human upper bound on session-level relation classification task 
are shown in Table \ref{tab:results}. The difficulties on session-level tasks is in proportional to the number of classes, as the performances of all of the models and human annotators decrease from 4-class task to 13-class task. The gap is about 10\% and 20\% on accuracy for models and human evaluation respectively. 

For Majority, the accuracy is even higher than a neural baseline (LSTM), while the F1-macro is the lowest due to the unbalanced data distribution between classes. The neural baselines mostly performs better than Random and Majority.

The comparison of performances on neural baselines is BERT$>$CNN$>$LSTM. LSTM is much weaker than CNN. The gap between them on 13-class classification is smaller than 4-class and 6-class classifications due to the fact them both of them failed on fine-grained classification task. The F1-macro is only 4.63\% and 9.20\% respectively with high variance.
BERT, a pre-trained language model baseline, performs much better and stable than the other two neural models with higher scores and lower variance. The gaps between evaluation metrics are also smaller than other baselines, indicating that it can handle the problem of imbalanced data to some extent.

The human performance is the average score of two annotators. We calculate the Cohen's Kappa between them, and the agreement are 0.429, 0.336 and 0.301 for 4-class, 6-class and 13-class relation classification tasks respectively. The agreement on 6-class and 13-class tasks are fair, and on more coarse-grained task, 4-class task, is moderate. It' s also quite difficult for human to identifying the relationship between interlocutors based on only one session. It seems that BERT outperforms human upper bound on accuracy of the 13-class classification task, but actually there is no significant differences between them, and F1-macro score of human upper bound is statistically significantly better than BERT with p-value less than 0.05.

\subsection{Pair-level Performance}

\label{sec:pair}
Table \ref{tab:results} also included the results on pair-level tasks. The difficulties between classification tasks on different granularity are the same as session-level classification tasks and the comparisons between baseline models are also the same: BERT$>$CNN$>$LSTM. 

By using the MRR metric to get the prediction on pair-level performance of each model as explained in Section \ref{sec:baselines}, it aggravate the polarization of the model performances. The strong baselines like CNN (except on 13-class classification task) and BERT achieve higher scores on pair-level tasks, while others, including LSTM and 13-class CNN model, performs even worse on pair-level tasks. The performance of LSTM is close to Majority baseline. The reason for this phenomenon is that we only assign one label for multiple sessions on pair-level tasks. If the model is weak, it tends to give some extremely unreasonable predictions on some of the sessions for a given pair of interlocutors. Even though there may be some correct predictions on session level, the final prediction for this pair is wrong. On the other hand, if the model is strong, it can give more reasonable predictions for most sessions. Then although there may be some wrong cases on session level, they will be tolerated. The performance of these models will increase.

The gap between CNN and BERT decreases from 4-class to 6-class tasks while increases greatly from 6-class to 13-class tasks. The convolution-based model seems more stable on coarse-grained tasks, and drops dramatically on the 13-class fine-grained task. On the contrary, the performance of fine-tuned language model decreases rapidly from 4-class to 6-class tasks and decreases slowly from 6-class to 13-class tasks. As a result, the advantage of BERT model on 6-class classification tasks is limited beyond the CNN baseline.

Human annotators also performance much better on pair-level tasks than session-level tasks. The Cohen's Kappa for two annotators are 0.698, 0.687 and 0.614 for 4-class, 6-class and 13-class classification tasks respectively, showing substantial agreements. The higher performances and agreement is consistent with the intuition that, with multiple sessions for a given pair, human are able to find more correlations between sessions and better understand the background of two interlocutors. In this way, we think pair-level relation classification tasks are more reasonable, challenging and meaningful for the development of current models. 

The gap between best baseline BERT and Human performances also showing the limitation of current models. We draw the confusion metrics for the best neural model BERT and human performances in Figure \ref{fig:confusion}. We can see that for coarse-level relation classification tasks, the performance of BERT and human has some similarities. They both did well on predicting the official relation type on 4-level task, and intimacy-peer relation type and official-peer relation type on 6-level task. For 13-classification tasks, BERT fails dramatically which may due to the unbalanced data distribution, tending to predict the relation type of ``{\em lovers}'', while human performance well on the relation type of ``{\em Workplace Superior-Subordinate}''.

\subsection{A First Step on Cross-session Consideration}
\label{sec:cross}

\begin{figure}[t!]
	\centering
	\includegraphics[width=0.7\columnwidth]{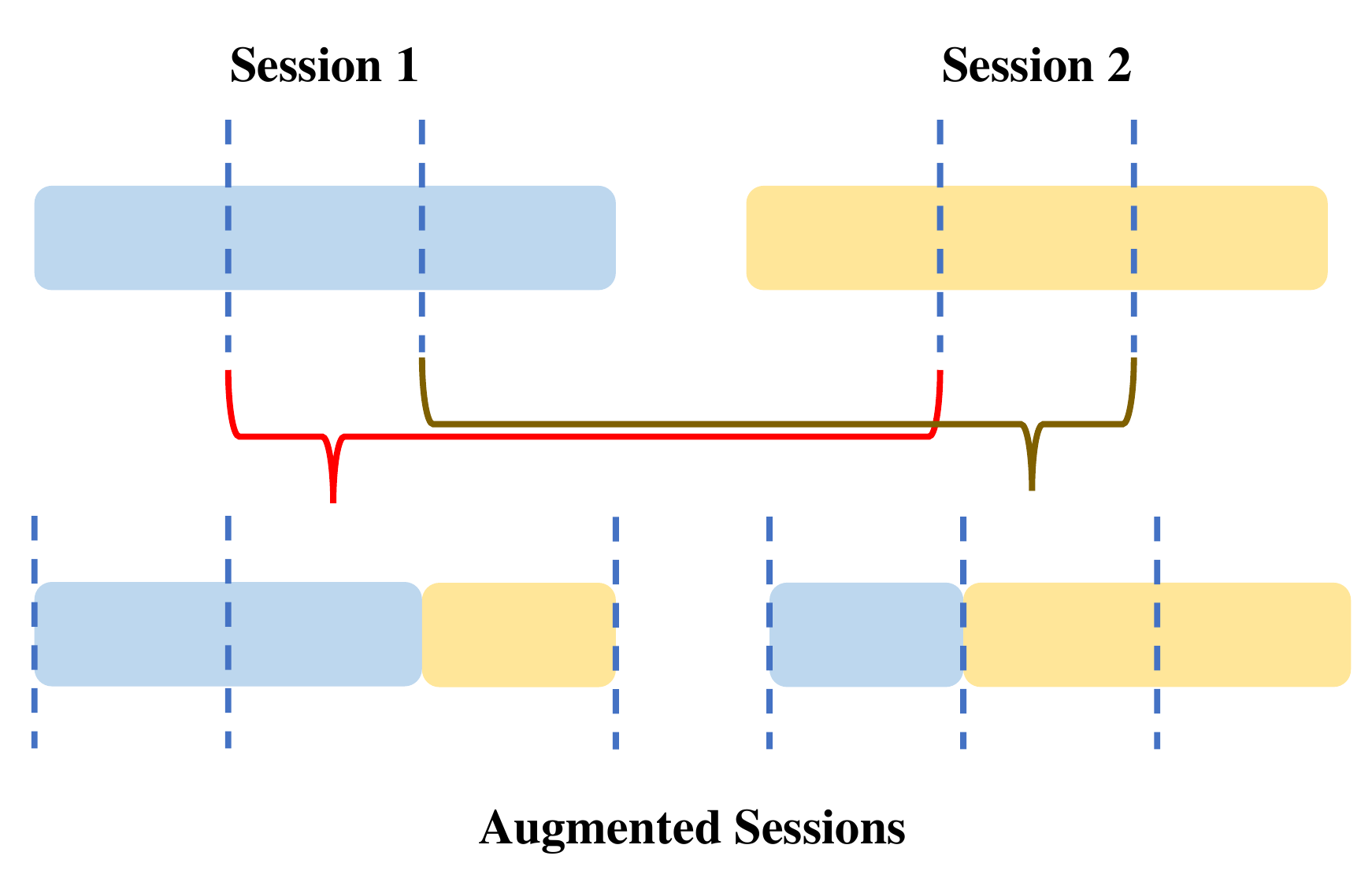}
	\caption{An illustration of data augmentation when $K=3$. }
	\label{fig:augment}
\end{figure}

For the pair-level classification tasks, our neural baselines give the final predictions by aggregating the predictions for each session. In this way, some interactions between sessions may be omitted. To clarify the existence of cross-session interaction in our RRDel dataset, we augment the original pair-level samples with multiple sessions as follows: i) Cut the session into $K$ pieces according to its length (the number of utterances). ii) Concatenate the session pieces at the same cut point in two consecutive sessions to generate a new session for the given pair. It should be noted that the order of sessions in each pair follows the chronological order. For example, in Figure \ref{fig:augment} is a pair with two sessions. Each session is cut into 3 pieces when $K=3$ and we get two augmented sessions for this pair.

\begin{table}[th]
	\centering
	\scriptsize
	\begin{tabular}{@{}lcccccc@{}}
		\toprule[1.5pt]
		& \multicolumn{2}{c}{\textbf{4-class}} & \multicolumn{2}{c}{\textbf{6-class}} & \multicolumn{2}{c}{\textbf{13-class}} \\ 
		& Acc & F1-macro                 & Acc & F1-macro                  & Acc & F1-macro \\
		\midrule
		$K=2$   &+1.73 &+2.50  &-0.40& -0.97& +2.13 & +0.63 \\		
		$K=3$    &+0.43 &+1.33 &+1.70 &+0.20 &+2.13 &+0.80 \\
		$K=4$    &-0.43 &+0.43 &+2.56 &+1.43 &+2.13 &+0.90 \\
		\bottomrule[1.5pt]
		
	\end{tabular}
	\caption{The pair-level classification results(\%) with data augmentation on the test set compared with BERT baseline. $K$ is the hyper-parameter for our proposed data augmentation method.}
	\label{tab:cross}
\end{table}

Using the best BERT model we trained above, we augmented the test set and re-evaluate the pair-level prediction results in Table \ref{tab:cross}. Most classification results are increased by augmentation operation. We further augmented all of the datasets in DDRel, and train and test the BERT baseline on the augmented dataset with $K=3$. The results on 6-class and 13-class enhanced significantly with accuracy equaling $49.13\%$ and $41.87\%$ respectively, and with F1-macro equaling $46.93\%$  and $26.83\%$ respectively. All of the results indicate the existence and importance of cross-session information for pair-level relationship classifications.

\subsection{Case Study and Future Directions}
\begin{figure}[t!]
	\centering
	\includegraphics[width=1.0\columnwidth]{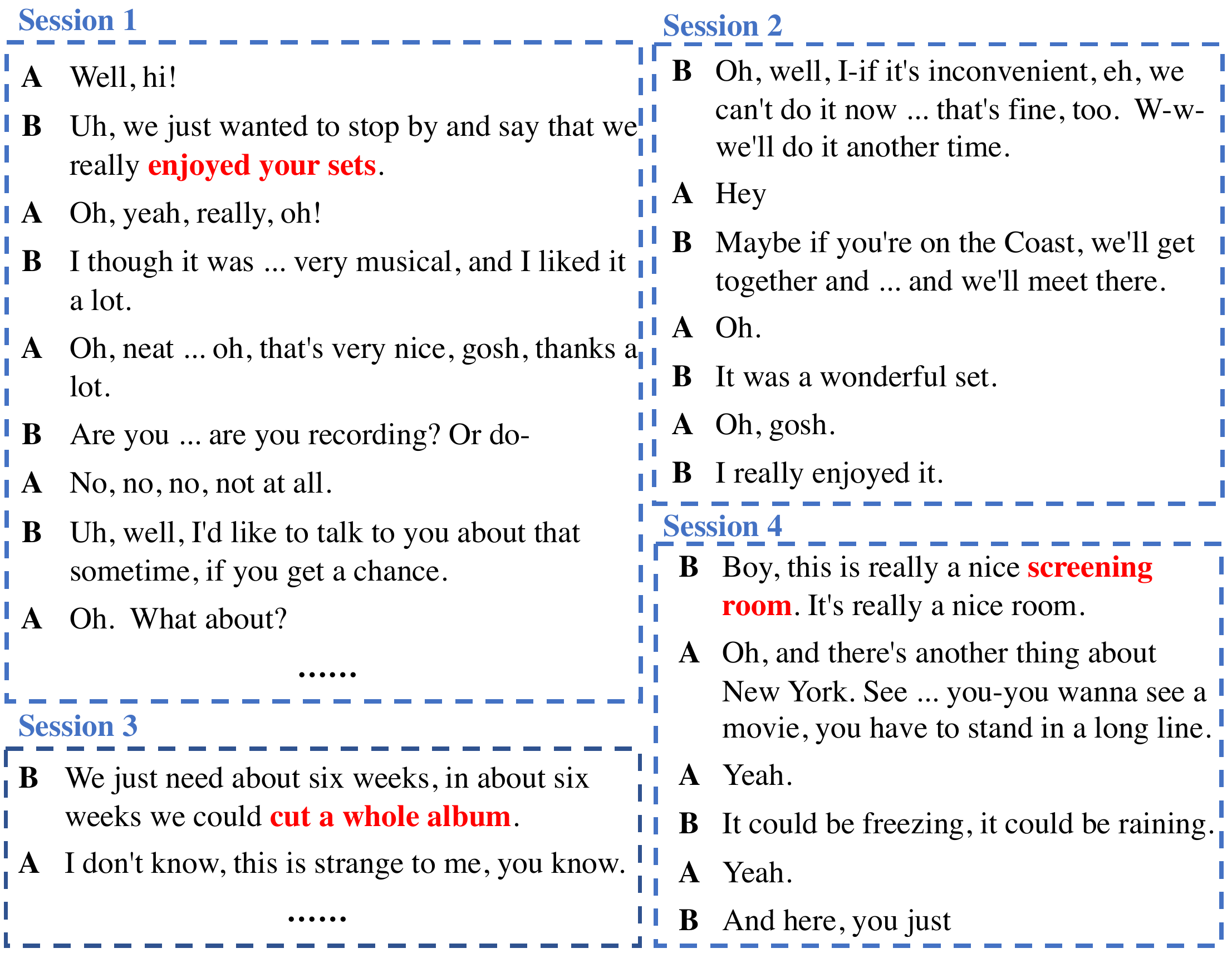}
	\caption{A pair-level case with 4 sessions. The words colored in red are possible classification cues. }
	\label{fig:case}
\end{figure}

We show a representative pair-level classification case with 4 sessions in Figure \ref{fig:case}. As for human, we can make inferences according to keywords 
such as ``enjoyed your sets'', ``cut a whole album'' and ``screening room''. Based on our background experience and knowledge, such conversations are more likely happening between two guys with cooperation in making music. The cues here are not obvious in single sessions but are very assuring when four sessions are considered together. Human choose ``Official'' while the best baseline BERT mistakes it to be ``Intimacy'' for this sample. The model may be confused by the informal expressions and emotional words such as ``wonderful''.

According to the case study, we consider the further research on this task as follows:

\textbf{Cross-session Consideration.} As talked about in Section \ref{sec:pair}, classification of a pair of interlocutors based on multiple sessions between them is a more reasonable and meaningful task. Due to the fact that the number of sessions for pairs varies a lot and it's difficult and unreasonable to concatenate all of the utterances in these sessions as the input for models, we only combines the prediction results of each session to get the final pair-level predictions and made a simple step on cross-session consideration by data augmentation. Developing models that could better find the cues between sessions is an important direction for current models.

\textbf{Commonsense Knowledge.} Another limitation of current models is due to the lack of commonsense knowledge, even for commonly pre-trained language models. Human can better inference the background of two interlocutors with the previous stories or experiences they have had. Further pre-training the language models on more similar corpus and incorporating the commonsense knowledge base, such as ConceptNet~\cite{SpeerCH17}, are possible solutions.

\section{Conclusion}

This paper proposes the interpersonal relation classification task for interlocutors in dyadic dialogues, accompanied with a new reasonable sized dialogue dataset called DDRel. The cross-session relation classification is raised for the first time and the results of baseline models shows the limitation for current methods on this new task. Models that taking advantages of multiple sessions and commonsense knowledge are expected to be explored as future work.

\bibliography{aaai2021}

\end{document}